\documentclass{article}

\usepackage[preprint]{neurips_2024}

\makeatletter
\renewcommand{\@notice}{}
\makeatother

\usepackage[utf8]{inputenc}
\usepackage[T1]{fontenc}
\usepackage{times}
\usepackage{latexsym}
\usepackage{lmodern}
\usepackage{microtype}
\usepackage{inconsolata}
\usepackage{textgreek}

\usepackage{amsmath}
\usepackage{amsfonts}
\usepackage{bm}
\usepackage{nicefrac}
\usepackage{amssymb}

\usepackage{graphicx}
\usepackage{booktabs}
\usepackage{multirow}
\usepackage{subcaption}
\usepackage{float}

\usepackage{xcolor}
\usepackage{tcolorbox}
\tcbuselibrary{breakable,skins}

\usepackage{pgfplots}
\pgfplotsset{compat=1.18}

\usepackage{tikz}
\usetikzlibrary{arrows.meta}

\usepackage{xspace}
\usepackage{tabularx}

\usepackage{url}
\usepackage{hyperref}
\hypersetup{hidelinks}

\usepackage{amsmath,amsfonts,bm}

\def\eqref#1{equation~\ref{#1}}

\def\1{\bm{1}}

\DeclareMathAlphabet{\mathsfit}{\encodingdefault}{\sfdefault}{m}{sl}
\SetMathAlphabet{\mathsfit}{bold}{\encodingdefault}{\sfdefault}{bx}{n}

\definecolor{color1}{HTML}{E74C3C}
\definecolor{color2}{HTML}{2ECC71}
\definecolor{color3}{HTML}{3498DB}
\definecolor{color4}{HTML}{F39C12}
\definecolor{color5}{HTML}{34495E}

\definecolor{color6}{HTML}{2ECC71}
\definecolor{color7}{HTML}{E74C3C}
\definecolor{color8}{HTML}{8B4513}

\definecolor{color9}{HTML}{E74C3C}
\definecolor{color10}{HTML}{2ECC71}
\definecolor{color11}{HTML}{8E44AD}

\definecolor{color12}{HTML}{1f77b4}
\definecolor{color13}{HTML}{ff7f0e}
\definecolor{color14}{HTML}{2ca02c}
\definecolor{color15}{HTML}{d62728}

\newcommand{\RoboCasa}{\textsf{RoboCasa}\xspace}

\title{Z-1: Efficient Reinforcement Learning for Vision-Language-Action Models}

\author{
Zioneer Robot Team
}

\begin{document}

\maketitle


\begin{abstract}
Vision-Language-Action (VLA) models offer a promising framework for robotic manipulation by connecting language instructions, visual observations, and continuous control.
However, most existing policies remain limited by behavior cloning or supervised fine-tuning (SFT) from fixed demonstrations, which provides limited opportunity to improve from the policy's own failures.
In this paper, we present Z-1, a reinforcement learning (RL) post-training framework for flow-based VLA models.
Built on top of $\pi_{0.5}$, Z-1 uses only publicly released RoboCasa demonstrations for SFT and then applies a task-wise Group Relative Policy Optimization (GRPO) strategy across $24$ standard RoboCasa tasks.
To improve the efficiency and stability of online optimization, Z-1 combines shared-prefix rollout construction, tree-structured trajectory branching, completion-aware reward calibration, and selective joint training of VLM and Action Expert.
Across all $24$ RoboCasa tasks, Z-1 achieves an average success rate of $80.6\%$, improving over its SFT initialization by $13.2\%$ points and outperforms the published sota models.
These results show that systematic GRPO post-training can substantially improve flow-based VLA policies without additional private demonstrations.
\end{abstract}

\section{Introduction}

Modern robotic manipulation increasingly requires policies that connect high-level language instructions~\citep{zhao2026surveylargelanguagemodels}, visual observations~\citep{bai2025qwen3vltechnicalreport}, and low-level continuous control~\citep{yang2025diffusionmodelscomprehensivesurvey}.
Vision-Language-Action (VLA) models~\citep{sapkota2026visionlanguageactionvlamodelsconcepts} provide a natural framework for this goal by unifying language understanding, visual perception, and action generation within a single policy.
Recent systems, including RT-2~\citep{brohan2023rt2visionlanguageactionmodelstransfer}, OpenVLA~\citep{kim2024openvlaopensourcevisionlanguageactionmodel}, $\pi_0$~\citep{black2026pi0visionlanguageactionflowmodel}, and $\pi_{0.5}$~\citep{intelligence2025pi05visionlanguageactionmodelopenworld} have demonstrated strong potential on complex manipulation tasks by leveraging semantic and visual priors from foundation models.

Despite this progress, most VLA policies are still primarily trained from human expert demonstrations through behavior cloning (BC)~\cite{torabi2018behavioralcloningobservation} or supervised fine-tuning (SFT)~\citet{brohan2023rt1roboticstransformerrealworld}.
Although imitation learning provides stable initialization, its performance is limited by the demonstration coverage~\citep{chi2024diffusionpolicyvisuomotorpolicy}. 
In long-horizon and compositional tasks, small prediction errors can accumulate and push the robot into states that are poorly represented in training data~\citep{kachaev2025dontblindvlaaligning}.
Moreover, BC and SFT provide no direct mechanism for improving from the policy's own failures, which limits its ability to optimize sparse task success through trial-and-error interaction.
These limitations motivate interaction-based post-training for VLA policies.

Reinforcement learning (RL) post-training offers a direct way to improve beyond imitation by collecting online rollouts and optimizing task-level rewards~\citep{lu2025vlarlmasterfulgeneralrobotic,li2025simplevlarlscalingvlatraining}. Recent studies have shown that RL can further improve the performance of SFT model ~\citep{chen2026pitextttrlonlinerlfinetuning,zhang2026reinflowfinetuningflowmatching,ye2025vlar1enhancingreasoningvisionlanguageaction,intelligence2025pi06vlalearnsexperience,intelligence2026pi07steerablegeneralistrobotic,pan2026sopscalableonlineposttraining}.
However, efficient and stable VLA-RL remains challenging.
Rollout generation is expensive for flow-based VLA models; sparse rewards and long horizons increase optimization variance; and freezing the vision-language backbone can limit the correction of failures caused by visual grounding, spatial reasoning, or language-conditioned perception.

In this work, we introduce Z-1, a modular RL post-training framework for flow-based VLA models.
Z-1 is built on top of the $\pi_{0.5}$ base model~\citep{intelligence2025pi05visionlanguageactionmodelopenworld} and uses only publicly released RoboCasa demonstration data for SFT~\citep{nvidia_robocasa_cosmos_policy_2024,nasiriany2024robocasalargescalesimulationeveryday}.
Our pipeline first performs per-scene SFT on RoboCasa demonstrations then applies a task-wise Group Relative Policy Optimization (GRPO)~\citep{shao2024deepseekmathpushinglimitsmathematical} post-training strategy across standard RoboCasa tasks.

Z-1 addresses these challenges with a modular GRPO design tailored to flow-based VLA policies.
It constructs rollout groups with shared prefixes and tree-structured branching to reduce redundant simulation and focus group-relative comparisons on task-critical interaction phases.
It further calibrates sparse rewards according to completion progress and selectively updates the vision-language module when action-only adaptation is insufficient. 
Together, these components improve the efficiency, stability, and adaptability of online post-training.

We evaluate Z-1 on $24$ standard RoboCasa tasks, and Z-1 improves substantially over its SFT initialization, achieving the best average success rate among the compared published RoboCasa results.
These results show that GRPO post-training can effectively improve flow-based VLA policies without additional private demonstrations.

Our contributions are summarized as follows:
\begin{itemize}
    \item We present Z-1, a modular GRPO post-training framework for flow-based VLA models, and we propose prefix-based rollout construction, including shared-prefix GRPO and tree-structured prefix branching, to reduce redundant rollout computation while preserving trainable trajectory segments.
    \item We develop success-aware reward decay and joint training of VLM and action expert for completion-aware reward calibration and perception-action co-adaptation.
    \item We achieve a new sota performance over RoboCasa benchmark under the public demonstration setting, with ablation analysis supporting key design choices.
\end{itemize}

\section{Related Work}

\subsection{Sota Models over RoboCasa}

Building on recent VLA systems such as RT-2~\citep{brohan2023rt2visionlanguageactionmodelstransfer}, OpenVLA~\citep{kim2024openvlaopensourcevisionlanguageactionmodel}, $\pi_0$~\citep{black2026pi0visionlanguageactionflowmodel}, and $\pi_{0.5}$~\citep{intelligence2025pi05visionlanguageactionmodelopenworld}, recent work has improved robotic action prediction by enriching visual-spatial representations, enlarging the training distribution, or building more generalist robot policies.

X-WAM~\citep{guo2026unified4dworldaction} explores world-action modeling for robotic manipulation by incorporating richer spatiotemporal visual structure into action prediction.
Related depth-aware VLA models show that explicit geometric information can improve spatial reasoning and manipulation performance~\citep{yuan2025depthvlaenhancingvisionlanguageactionmodels}.
These methods highlight the importance of structured visual-spatial representations for action generation.

Another line of work studies generalist robot foundation models trained across diverse embodiments, environments, and tasks.
GR00T~\citep{nvidia2025gr00tn1openfoundation} is a representative open foundation model for generalist humanoid robots, trained to map multimodal observations and language instructions to robot actions across diverse skills.
Such models often improve performance by increasing model capacity, data diversity, or embodiment coverage, while recent studies on robot data curation further show that data quality and coverage are critical for scalable robot learning~\citep{hejna2025robotdatacurationmutual}.

Z-1 is orthogonal to these directions.
Rather than introducing a new world model, geometric representation, or generalist architecture, Z-1 studies how an existing flow-based VLA policy can be improved through task-level online reinforcement learning.
This focus allows us to study the effect of RL post-training algorithm in a setting where only publicly released demonstrations are available.
In particular, Z-1 investigates whether rollout construction, reward calibration, and trainable-module selection can substantially improve a flow-based VLA policy without relying on additional private demonstrations.

\subsection{RL Post-Training for VLA Models}

Reinforcement learning post-training has recently emerged as a promising approach of improving VLA policies beyond demonstration-based initialization.
In contrast to behavior cloning or supervised fine-tuning, online RL allows the policy to collect its own interaction data and optimize task-level success signals.
This is especially important for long-horizon manipulation tasks, where small imitation errors can compound and lead to states that are poorly covered by demonstrations.

Several recent studies have explored this direction.
Recent studies, including $\pi$RL~\citep{chen2026pitextttrlonlinerlfinetuning}, ReinFlow~\citep{zhang2026reinflowfinetuningflowmatching}, and VLA-R1~\citep{ye2025vlar1enhancingreasoningvisionlanguageaction}, have explored online RL fine-tuning after SFT, flow-based policy fine-tuning, and reasoning-oriented post-training for VLA policies.
Recent $\pi$-series systems, including $\pi^*_{0.6}$~\citep{intelligence2025pi06vlalearnsexperience} and $\pi_{0.7}$~\citep{intelligence2026pi07steerablegeneralistrobotic}, further study learning from robot experience and steerable generalist robot policies, while SOP~\citep{pan2026sopscalableonlineposttraining} investigates scalable online post-training.

Despite this progress, efficient and stable RL post-training for flow-based VLA models remains challenging.
First, online interaction remains costly in robot RL, motivating simulation- and digital-twin-based training pipelines~\citep{xu2026twinrldigitaltwindrivenreinforcement}; this cost is amplified when each GRPO group member executes a full long-horizon trajectory independently.
Second, sparse success rewards provide limited credit assignment, and binary task outcomes may fail to distinguish different successful trajectories.
Third, freezing the vision-language backbone improves stability but can limit adaptation when failures arise from visual grounding, spatial reasoning, or language-conditioned perception rather than only low-level control.
Recent spatially guided VLA models also highlight the importance of explicit spatial grounding for action prediction~\citep{chen2025internvlam1spatiallyguidedvisionlanguageaction}.

Z-1 aims to mitigate these challenges with a modular GRPO post-training framework.
Instead of treating each rollout as an independent full trajectory, Z-1 constructs rollout groups with shared prefixes and tree-structured branching to reduce redundant prefix execution and focus comparisons on task-critical segments.
It further introduces completion-aware reward calibration through Success-Aware Reward Decay and selectively updates the vision-language module together with the action expert for tasks where AE-only adaptation is insufficient.
Together, these design choices emphasize rollout efficiency, optimization stability, and representation-action co-adaptation for flow-based VLA post-training.

\section{Methodology}
\label{sec:method}

Z-1 follows a two-stage training pipeline.
The first stage performs per-scene supervised fine-tuning (SFT), and the second stage applies task-wise GRPO post-training.
Starting from a pretrained $\pi_{0.5}$ base model, we first adapt Z-1 to RoboCasa using public demonstrations and then applies online interaction-based refinement to tasks selected by pre-evaluation training diagnostics.
For the reinforcement learning stage, each task uses a fixed subset of Z-1 modules, chosen before final evaluation based on training-stage diagnostics such as SFT performance, early AE-only GRPO progress, and failure modes observed in training rollouts.
No final evaluation rollouts are used for post-training or module selection.
The available modules include Shared-Prefix GRPO, Tree-Structured Prefix Branching, Success-Aware Reward Decay, and VLM--Action Expert joint training.

\subsection{Two-Stage Training}

Z-1 builds on the pretrained $\pi_{0.5}$ policy and contains two stages.

\subsubsection{Stage 1: Per-scene SFT}
We first adapt $\pi_{0.5}$ to RoboCasa using publicly released demonstrations.
Due to the limited amount of data, we train one SFT model per scene category rather than one model per task.
Tasks within the same scene share object layouts, visual contexts, and manipulation primitives, so per-scene SFT improves data efficiency while avoiding per-task data fragmentation.
During SFT, both the vision-language module and the action expert are updated.
For each scene category, we select a stable SFT checkpoint with a held-out validation loss around 0.01 as the task-wise initialization.
Depending on the scene category, we either directly fine-tune on the target scene or use an all-data warm-up followed by scene-specific fine-tuning.
Detailed SFT schedules are provided in Appendix~\ref{app:training_details}.

\subsubsection{Stage 2: Per-task GRPO post-training}
Starting from the corresponding scene-specific SFT checkpoint, we perform task-wise GRPO post-training for tasks selected by training-stage diagnostics.
When GRPO is applied, optimizing each task separately avoids gradient interference across heterogeneous tasks and allows each task to use a fixed subset of Z-1 modules.
The enabled modules are not mutually exclusive.
That is, a task may combine prefix-based rollout construction, completion-aware reward calibration, and VLM--Action Expert joint training when indicated by training diagnostics.
For example, tasks with reliable approach behavior can enable flat Shared-Prefix GRPO for lightweight prefix reuse, whereas tasks requiring approach-phase refinement can enable Tree-Structured Prefix Branching to introduce rollout diversity progressively within the prefix.
Tasks with large variation in successful completion time can enable Success-Aware Reward Decay, and tasks where AE-only GRPO is insufficient can further enable VLM--Action Expert joint training.

Formally, we denote $\pi_{\theta}$ as the base model initialized by SFT.
For each training instance, GRPO samples a group of $G$ rollouts $\mathcal{G}=\{\tau^{(i)}\}_{i=1}^{G}$ with returns $R_i$.
The group-relative advantage is computed as

\begin{equation}
A_i
=
\dfrac{R_i - \mu_R}{\sigma_R + \epsilon},
\end{equation}
where

\begin{equation}
\mu_R
=
\dfrac{1}{G}\sum_{j=1}^{G}R_j,
\quad
\sigma_R
=
\sqrt{
\dfrac{1}{G}\sum_{j=1}^{G}
\left(R_j-\mu_R\right)^2
}.
\end{equation}
Z-1 optimizes the clipped group-relative objective over trainable action chunks:
\begin{equation}
\mathcal{L}_{\mathrm{GRPO}}(\theta)
=
-
\frac{1}{\sum_{i=1}^{G}|\mathcal{B}_i|}
\sum_{i=1}^{G}
\sum_{(x,a)\in\mathcal{B}_i}
\min
\left(
\rho_{\theta}(x,a) A_i,
\mathrm{clip}
\left(
\rho_{\theta}(x,a),
1-\eta,
1+\eta
\right) A_i
\right),
\end{equation}
where
\begin{equation}
\rho_{\theta}(x,a)
=
\frac{
\pi_{\theta}(a\mid x)
}{
\pi_{\theta_{\mathrm{old}}}(a\mid x)
}.
\end{equation}
Here, $\mathcal{B}_i$ denotes the set of trainable action chunks in rollout $i$.
For flow-based action generation, we follow the flow-SDE formulation of $\pi$RL~\citep{chen2026pitextttrlonlinerlfinetuning}.
This converts the deterministic flow sampling path into a stochastic Markov process with tractable transition likelihoods, allowing the action-chunk log-probability and the GRPO likelihood ratio to be computed from the rollout-time SDE transitions.
Details are provided in Appendix~\ref{app:flow_sde_ratio}.
Prefix chunks masked by Shared-Prefix GRPO are excluded from $\mathcal{B}_i$, and we do not use an additional KL regularizer.

\subsection{Shared-Prefix GRPO}

Many manipulation trajectories usually contain two phases, an approach phase where the end effector approaches the objects and a second fine-manipulation phase, such as grasping.
Standard GRPO is designed to compare final rewards of full trajectories, each sampled independently, during one group.
However, the approach phases of trajectories have different intermediate states, and the final reward may reflect the sample quality of approach phase rather than the manipulation behavior, which should receive more attention during reinforcement learning.
Thus, standard GRPO might dilute the group-relative reward signal, especially when task-critical decisions occur near the interaction region.

To tackle this, we propose Shared-Prefix GRPO to reduce this confounding by forcing rollouts in the same GRPO group to share a common prefix.
In detail, for each group, a leader rollout executes the prefix over the action-chunk window $[0,P)$, where $P$ is the prefix length in action chunks.
The simulator state at the branching point is then cloned to initialize the remaining group trajectories.
All trajectories then execute independent suffixes after the branching point, so their returns compare behaviors starting from the same entering state.

The flat shared-prefix variant uses this one-shot branching construction.
Prefix chunks are treated as a shared condition and excluded from the policy-gradient loss, and only suffix chunks enter $\mathcal{B}_i$ in the GRPO objective.
Because the shared prefix is used only to define a common branching state, masking prefix chunks avoids assigning different suffix returns to identical prefix actions.
This variant reduces redundant prefix computation and is most useful when the approach phase is already reliable after SFT.
However, it is conservative because the entire prefix window is masked.
For tasks whose approach behavior still requires RL refinement, Z-1 uses Tree-Structured Prefix Branching to reduce the fully masked segment.

\subsection{Tree-Structured Prefix Branching}

Tree-Structured Prefix Branching~\citep{cao2026treeadvtreestructuredadvantageredistribution} generalizes flat Shared-Prefix GRPO by replacing one-shot branching with progressive hierarchical branching.
Instead of keeping the entire group identical until the final branching point, it introduces intermediate branching points within the prefix.
This creates a tree of rollout clusters, where early segments are shared by more rollouts, while later segments gradually become more diverse.

More specifically, only the fully shared root segment is masked.
Partially shared segments remain trainable because different rollout clusters have already branched and can receive different rewards and learning signals.
Partially shared chunks are included only after their cluster has separated from other clusters, so their learning signal is associated with branch-specific returns rather than with an identical full-group prefix.
Compared with flat Shared-Prefix GRPO, this reduces over-masking and allows more approach-phase chunks to contribute to the GRPO update.
The GRPO objective and advantage estimator remain unchanged, and only the rollout group construction differs.
Because Tree-Structured Prefix Branching uses a smaller group size in our implementation, it can reduce the number of effective GRPO groups that provide non-zero learning signals under a fixed batch size.
We therefore increase the training batch size for this variant so that the number of effective learning groups remains comparable to the standard GRPO setting.
We provide the detailed branching schedule, action-recording procedure, and implementation details in Appendix~\ref{app:tree_prefix_details}.

\subsection{Success-Aware Reward Decay}

Manipulation tasks often provide sparse success rewards: a rollout receives a positive score only when the task is completed.
Treating all successful rollouts identically makes the reward signal coarse, since an early completion and a late completion receive the same score.

To address this problem, we use \textbf{Success-Aware Reward Decay} to introduce an ordering among successful rollouts according to completion time.
The decay is applied only to successful trajectories.
Let $r$ be the original success score, and let $d_i$ denote the number of action chunks elapsed from the start of rollout $i$ until its first successful completion.
We define the calibrated score as
$
\tilde{r}_i
=
r \cdot \gamma^{d_i},
$
where $\gamma \in (0,1]$ is the reward decay factor.
When $\gamma=1$, the original sparse reward is unchanged; when $\gamma<1$, earlier successful completions receive larger positive scores than later successful completions.
Failed rollouts keep their original failure reward.

The calibrated score is used as the rollout return in GRPO.
For groups containing both successful and failed rollouts, standard group-relative normalization is used, so the success-failure distinction remains dominant.
For groups where all rollouts succeed, Z-1 uses a non-negative completion-time ordering:
$
A_i
=
R_i
-
\min_j R_j.
$
This replaces the standard mean-normalized advantage only for all-success groups.
Thus, the slowest successful rollout receives zero advantage, while earlier successful completions receive positive advantages.
This avoids treating slower successful rollouts as negative examples, which can otherwise over-emphasize trajectory shortening at the expense of reliable task completion.
This preserves the sparse success objective while providing finer group-relative feedback without task-specific dense reward shaping, an additional reward model, or explicit trajectory-length penalties.

\textbf{Selective VLM--Action Expert Joint Training} (Sel-JT)

Many VLA-RL pipelines freeze the vision-language backbone and update only the action expert during online post-training.
This AE-only setting is efficient and stable when the SFT checkpoint already provides sufficient visual grounding and language-conditioned representations.
We also apply the AE-only GRPO as the default configuration.

However, AE-only adaptation can be insufficient for tasks with low SFT success or perception-sensitive failures, where the limiting factor may be inaccurate object localization, weak spatial grounding, or poor alignment between vision-language representations and action generation.
For such tasks, Z-1 selectively enables \textbf{VLM--Action Expert joint training} according to training-stage diagnostics, including SFT success, early AE-only GRPO progress, and failure modes observed in training rollouts.
The selected trainable module configuration is fixed before final evaluation, and no final evaluation rollouts are used for module or model selection.

Both AE-only and joint-training settings use the same rollout construction, advantage estimator, likelihood-ratio computation, and clipped GRPO objective.
They differ only in which model parameters receive gradients.
In the AE-only setting, GRPO freezes the PaliGemma vision-language backbone and updates only the action expert.
When joint training is enabled, the vision-language backbone is included in the trainable parameter set and optimized together with the action expert.
This allows task-level RL feedback to adapt perception, grounding, and action generation jointly without changing the underlying GRPO objective.
Implementation details are provided in Appendix~\ref{app:selective_joint_training}.

\section{Experiments}

We evaluate Z-1 on \RoboCasa, a large-scale simulated household manipulation benchmark. 
Our evaluation covers $24$ standard manipulation tasks, including spanning doors, drawers, pick-and-place, microwave, sink/faucet, stove, and coffee-making scenarios.
We use task success rate as the primary evaluation metric.
All success rates are reported in percentages and rounded to one decimal place unless otherwise specified.

The experiments are designed to answer three research questions:
\begin{enumerate}
\item How Z-1 compares with previously reported RoboCasa SOTA,
\item To what extent GRPO post-training improves the supervised fine-tuned initialization, and 
\item To what extent different task categories can benefit from online post-training.
\end{enumerate}

\subsection{Experimental Setup}

\subsubsection{Base model and training pipeline}
As we investigate a post-training method, we build Z-1 on top of the pretrained \ensuremath{\pi_{0.5}} base model.
As described in Section~\ref{sec:method}, we first perform supervised fine-tuning (SFT) using the publicly released RoboCasa demonstrations, and then we apply task-specific GRPO post-training when selected by training-stage diagnostics.
The SFT stage uses $1,199$ demonstrations in total and is implemented with the official OpenPI JAX training code.

We use scene-specific SFT as the default adaptation strategy.
For the door, drawer, and microwave scenes with limited data, we adopt a two-stage SFT schedule to provide a stronger initialization, where the model is first fine-tuned on all $1,199$ RoboCasa demonstrations and then further fine-tuned on the target scene-specific demonstrations.
For the other scenes, we directly fine-tune on the corresponding scene-specific demonstrations.
This produces the per-scene SFT initialization used by the post-training pipeline.
We denote these models as Z-1 SFT and denote the final task-wise policy selected by the Z-1 post-training pipeline as Z-1 RL.

\subsubsection{RL implementation}
Our GRPO implementation is based on the RLinf framework and operates on the Z-1 SFT checkpoints.
For tasks selected for RL, post-training starts from the corresponding scene-specific SFT checkpoint and collects online rollouts in the RoboCasa simulator.
We implement the Z-1 modules described in Section~\ref{sec:method} on top of the original RLinf codebase, including Shared-Prefix GRPO, Tree-Structured Prefix Branching, Success-Aware Reward Decay, and selective VLM--Action Expert joint training. 
These modules modify rollout construction, reward calibration, and trainable parameter selection while keeping the underlying group-relative policy optimization objective unchanged.

Unless otherwise specified, RL updates only the action expert.
When training-stage diagnostics indicate that action-expert-only adaptation is insufficient, we enable joint training of the vision-language module and the action expert.
We do not use an additional KL regularizer during GRPO post-training. 
For each task, the module configuration is fixed before the final evaluation.

\subsubsection{Evaluation protocol}
We evaluate Z-1 SFT and Z-1 RL using independent evaluation rollouts with multiple random seeds, without reusing any training trajectories.
Task success rate is computed over all evaluation rollouts.
Category-level results are computed as the unweighted average of task-level success rates within each category, and the overall average is computed as the unweighted average across all $24$ tasks.
Detailed evaluation budgets, including the number of seeds and rollouts per task, are provided in Appendix~\ref{app:evaluation_protocol}.

\subsubsection{Baselines and Comparisons}
We compare Z-1 with four previously published RoboCasa results: GR00T, GR00T N1.5, Video Generators are Robot Policies (Video Policy), and X-WAM.
The GR00T and GR00T N1.5 results are taken from their reported RoboCasa evaluations~\citep{nvidia2025gr00tn1openfoundation}; the Video Policy results are taken from the reported RoboCasa evaluation in~\citep{liang2025videogeneratorsrobotpolicies}; and the X-WAM results are taken from the reported RoboCasa evaluation in~\citep{guo2026unified4dworldaction}.
We compare against the numbers reported in the corresponding papers and interpret the comparison as a reference to published RoboCasa performance rather than a fully controlled reproduction study.
For consistency, all baseline and Z-1 results are reported as success rates in percentages.

\subsection{Main Results}

\begin{figure*}[t]
\centering
\begin{tikzpicture}
\begin{axis}[
    ybar,
    width=0.99\textwidth,
    height=6.7cm,
    bar width=3.8pt,
    enlarge x limits=0.10,
    ymin=10,
    ymax=106,
    ylabel={Success Rate (\%)},
    symbolic x coords={Door,Drawer,PnP,Micro.,Sink,Stove,Coffee,Avg.},
    xtick=data,
    ytick={20,40,60,80,100},
    x tick label style={
        font=\small,
        yshift=-1pt
    },
    y tick label style={font=\small},
    ymajorgrids=true,
    grid style={dashed,gray!30},
    axis line style={gray!60},
    tick style={gray!60},
    label style={font=\small},
    clip=false,
    legend style={
        at={(0.5,1.10)},
        anchor=south,
        legend columns=6,
        draw=none,
        font=\small,
        /tikz/every even column/.append style={column sep=0.14cm}
    },
    legend image code/.code={
        \draw[#1] (0cm,-0.07cm) rectangle (0.18cm,0.07cm);
    },
]

\addplot[fill=teal!35, draw=teal!80!black] coordinates {
    (Door,58.0)
    (Drawer,89.0)
    (PnP,22.6)
    (Micro.,74.5)
    (Sink,63.0)
    (Stove,41.5)
    (Coffee,60.3)
    (Avg.,49.7)
};

\addplot[fill=cyan!35, draw=cyan!70!black] coordinates {
    (Door,75.2)
    (Drawer,90.5)
    (PnP,39.5)
    (Micro.,73.5)
    (Sink,76.3)
    (Stove,30.0)
    (Coffee,66.3)
    (Avg.,59.7)
};

\addplot[fill=purple!30, draw=purple!75!black] coordinates {
    (Door,90.5)
    (Drawer,71.0)
    (PnP,50.5)
    (Micro.,91.0)
    (Sink,67.3)
    (Stove,18.0)
    (Coffee,63.3)
    (Avg.,63.3)
};

\addplot[fill=orange!35, draw=orange!85!black] coordinates {
    (Door,93.2)
    (Drawer,92.5)
    (PnP,71.5)
    (Micro.,87.5)
    (Sink,86.0)
    (Stove,57.5)
    (Coffee,74.3)
    (Avg.,79.2)
};

\addplot[fill=blue!40, draw=blue!80!black] coordinates {
    (Door,93.2)
    (Drawer,83.4)
    (PnP,53.9)
    (Micro.,94.8)
    (Sink,63.2)
    (Stove,29.2)
    (Coffee,69.8)
    (Avg.,67.4)
};

\addplot[fill=red!65, draw=red!90!black] coordinates {
    (Door,97.0)
    (Drawer,96.1)
    (PnP,70.9)
    (Micro.,94.8)
    (Sink,94.3)
    (Stove,44.5)
    (Coffee,75.2)
    (Avg.,80.6)
};

\legend{GR00T, GR00T N1.5, Video Policy, X-WAM, Z-1 SFT, Z-1 RL}

\node[
    font=\scriptsize\bfseries,
    text=red!80!black,
    align=center,
    fill=white,
    fill opacity=0.92,
    text opacity=1,
    inner sep=1.4pt,
    rounded corners=1pt
] at ([xshift=0pt]axis cs:Avg.,90.8) {+1.4 pp\\over X-WAM};

\draw[
    -{Latex[length=1.5mm]},
    red!80!black,
    line width=0.6pt
] ([xshift=4pt]axis cs:Avg.,86.8) -- ([xshift=10pt]axis cs:Avg.,80.0);

\draw[black, line width=0.5pt]
    (rel axis cs:-0.015,0.012) -- (rel axis cs:0.015,0.045);

\draw[black, line width=0.5pt]
    (rel axis cs:-0.015,0.045) -- (rel axis cs:0.015,0.078);

\end{axis}
\end{tikzpicture}
\caption{
Category-level and averaged success rates on RoboCasa across all compared methods.
PnP and Micro. denote Pick-and-Place and Microwave, respectively.
The Avg. group reports the mean success rate over all $24$ RoboCasa tasks.
Baseline results are taken from the corresponding papers, while Z-1 SFT and Z-1 RL are evaluated by us.
Z-1 RL achieves the best average performance among the compared published methods, with an average success rate of 80.6\%, improving over X-WAM's $79.2\%$ by $1.4$ percentage points.
}
\label{fig:robocasa_category_all}
\end{figure*}
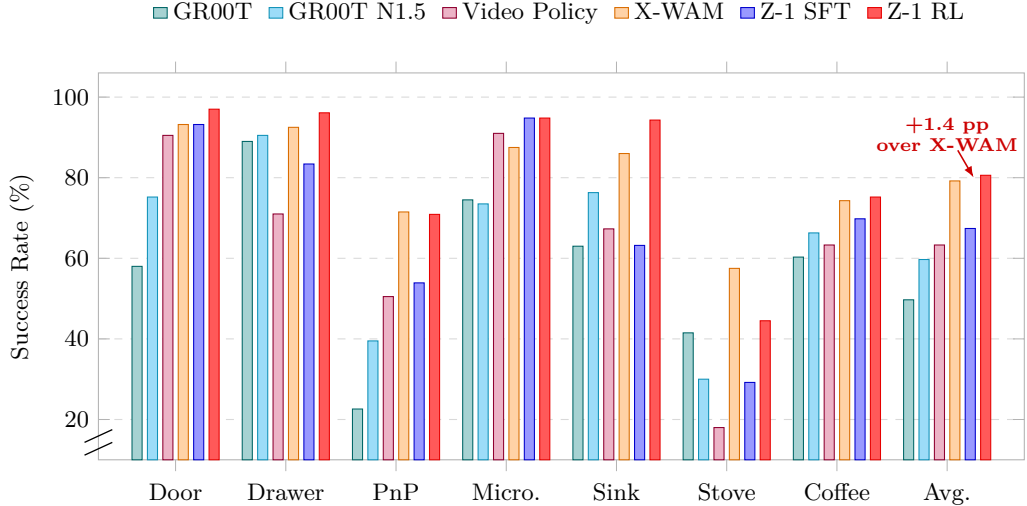

Figure~\ref{fig:robocasa_category_all} summarizes the category-level results of Z-1 with and compared baselines on \RoboCasa.
Z-1 RL achieves a sota average success rate of $80.6\%$ across $24$ RoboCasa tasks under the public RoboCasa demonstration setting.
Compared with its Z-1 SFT initialization, Z-1 RL improves the average success rate from $67.4\%$ to $80.6\%$, corresponding to a gain of $13.2$ percentage points.
Also, when compared across $7$ categories, Z-1 RL consistently improved Z-1 SFT's performance.
These demonstrate that the Z-1 RL post-training method provides substantial gains beyond supervised fine-tuning.

Compared the averaged success rate with other sota baselines, Z-1 RL outperforms the reported GR00T, GR00T N1.5, Video Policy, and X-WAM results by $30.9$, $20.9$, $17.3$, and $1.4$ percentage points, respectively.
Although the average gain over X-WAM, the strongest published baseline in our comparison, is modest, Z-1 RL outperforms X-WAM across $5$ categories out of $7$ in total.
Note that X-WAM introduces the depth information, while our Z-1 RL only uses publicly released RoboCasa demonstrations.
The improvement is especially pronounced in drawer, sink/faucet, and several pick-and-place tasks, where online RL can refine interaction behavior beyond the SFT policy.

\subsubsection{RL Improvements over Different Tasks}
The gains from Z-1 RL are not uniform across categories. 
We observe that RL improve from $93.2\%$ to $97.0\%$ on door tasks, from $83.4\%$ to $96.1\%$ on drawer tasks, and improves from $63.2\%$ to $94.3\%$ on sink/faucet tasks. 
These categories often require precise interaction after the robot reaches the target object, which aligns with the motivation of Shared-Prefix GRPO and Tree-Structured Prefix Branching.

We also observe substantial improvement on pick-and-place tasks, from $53.9\%$ to $70.9\%$ specifically. 
Microwave tasks already benefit from the SFT stage and remain at $94.8\%$ after RL, suggesting that the SFT checkpoint already captures the required behavior for this category. 
Stove tasks remain challenging that the success rates of all methods are low.
Z-1 RL improves over Z-1 SFT, but still underperforms X-WAM on this category. 
This indicates that stove manipulation may require stronger representation adaptation, more targeted exploration, or additional failure recovery mechanisms.

\subsection{Ablation and Analysis}

\subsubsection{Effect of Shared Prefix and Prefix Tree}

\begin{figure*}[t]
    \centering
    \includegraphics[width=0.95\textwidth]{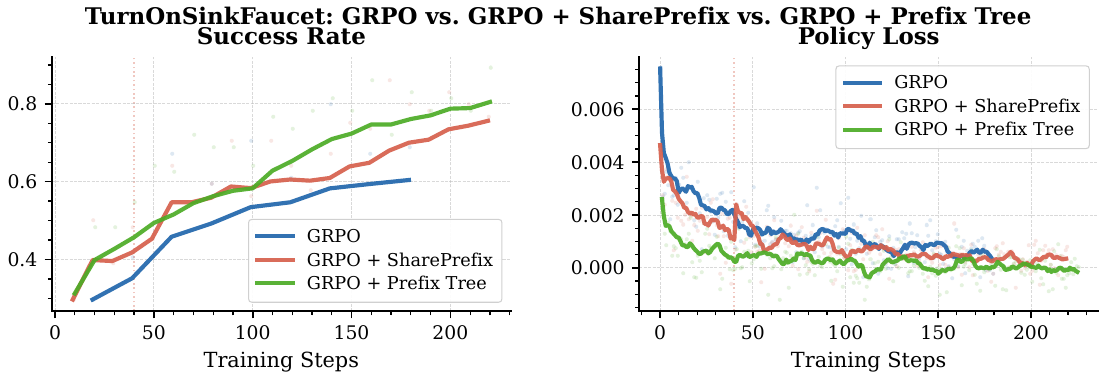}
    \caption{
        Training performance comparison on the TurnOnSinkFaucet task across three GRPO variants:
        vanilla GRPO, GRPO with a flat shared prefix (SharePrefix), and GRPO with Tree-Structured Prefix Branching (Prefix Tree).
        Left: success rate over training steps.
        Right: policy loss over training steps.
        Both prefix-based variants improve over vanilla GRPO.
        Prefix Tree achieves the strongest final success rate and a faster loss decrease in this run,
        while flat SharePrefix provides a more stable alternative in practice.
    }
    \label{fig:shared_prefix}
\end{figure*}

We analyze the effect of prefix-based rollout construction on the \textbf{TurnOnSinkFaucet} task.
This task requires the policy to first reach the sink area and then execute precise contact-rich manipulation to turn on the faucet.
It therefore provides a useful setting for studying whether group-relative policy optimization benefits from reducing variation in the early trajectory segment.

Figure~\ref{fig:shared_prefix} compares three variants, vanilla GRPO, flat Shared-Prefix GRPO, and Tree-Structured Prefix Branching.
Overall, both prefix-based variants achieve higher success rates than vanilla GRPO.
This result suggests that sharing early trajectory prefixes can make group-relative comparisons more informative.
Without prefix sharing, rollouts in the same group may differ substantially during the approach phase, such as in the arm pose or the state from which the faucet interaction begins.
These differences can introduce irrelevant variation into the reward comparison and make it harder to assign credit to the generated actions that determine task success.

Flat Shared-Prefix GRPO addresses this issue by branching multiple suffix rollouts from the same prefix state.
As a result, reward differences within a group are more directly tied to post-prefix manipulation behavior.
Compared with vanilla GRPO, this yields a cleaner optimization signal and leads to more stable improvement in success rate.

Prefix Tree further improves the final success rate in this experiment. 
It also shows a faster policy-loss decrease, indicating that preserving more trainable trajectory segments can provide a stronger learning signal. 
Unlike flat Shared-Prefix GRPO, which masks the full shared prefix, Prefix Tree only masks the fully shared root segment and allows later partially branched segments to contribute to the policy update. 
This method can reduce over-masking and make better use of rollout data.

However, Prefix Tree introduces a stability trade-off. 
Because more prefix-region actions are exposed to policy updates, the resulting optimization signal is less conservative and can be more sensitive to task configuration, prefix length, and rollout stochasticity. 
In our experiments, this sometimes led to unstable training dynamics. 
By contrast, flat Shared-Prefix GRPO is more conservative as it masks the full shared prefix and focuses the update on the suffix after the shared state.
Although this may discard some trainable prefix-region actions, it provides a robust and practical alternative when training stability is the primary concern.

Overall, these results show that prefix-based rollout construction improves GRPO training on long-horizon manipulation tasks. 
Prefix Tree can achieve stronger peak performance when training remains stable, while flat Shared-Prefix GRPO offers a better stability--performance trade-off in practice.

\begin{figure*}[t]
    \centering
    \includegraphics[width=\textwidth]{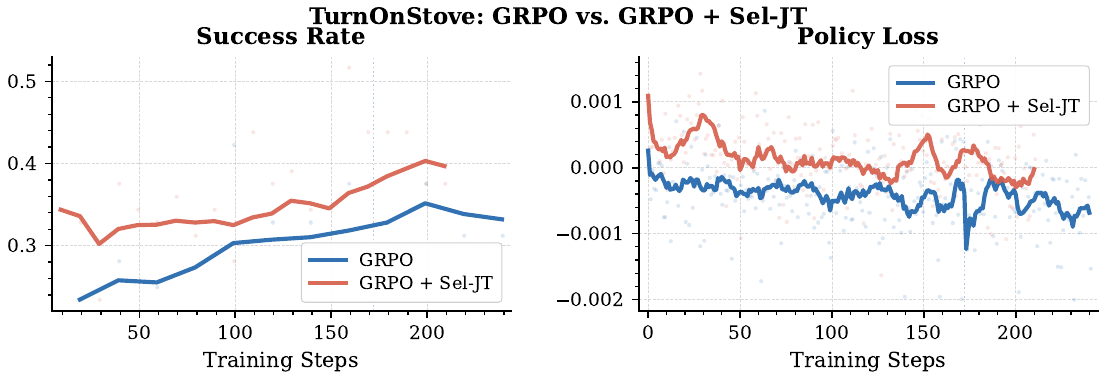}
    \caption{
        Training curves on the \textbf{TurnOnStove} task.
        \textit{Left}: Success rate over training steps.
        \textit{Right}: Policy loss over training steps.
        Our method (\textbf{GRPO + Sel-JT}, red) consistently achieves higher success rate
        and lower policy loss compared to the baseline (\textbf{GRPO}, blue),
        demonstrating the effectiveness of Selective VLM--Action Expert Joint Training.
    }
    \label{fig:turnonStove_metrics}
\end{figure*}

\subsubsection{Effect of Selective VLM--Action Expert Joint Training}

We further evaluate the effect of Selective VLM--Action Expert Joint Training on the challenging TurnOnStove task. 
This task requires accurate visual grounding of the stove knob, precise spatial reaching, and contact-rich manipulation, making it difficult to solve by only adapting the action expert when the visual-language representation is suboptimal.

Figure~\ref{fig:turnonStove_metrics} compares standard GRPO, which updates only the action expert, with our selective joint-training variant, denoted as GRPO + Sel-JT. 
GRPO + Sel-JT consistently achieves a higher success rate than the AE-only baseline throughout the training phase, indicating that updating the VLM together with the action expert provides a more effective learning signal for this difficult task. 
In contrast, AE-only GRPO is constrained by the frozen representation and shows limited improvement.

The policy-loss curve further supports these observations.
GRPO + Sel-JT obtains a lower policy loss than AE-only GRPO, suggesting that joint training makes the policy easier to optimize under task-level RL feedback.
This is consistent with our motivation that for tasks where failures may arise from inaccurate visual grounding or weak language-conditioned spatial representation, freezing the VLM limits the action expert to operate on imperfect features.
Selective joint training allows the model to adapt perception, grounding, and action generation together, leading to more stable and effective improvement.

These results validate the selective nature of our method. 
Rather than jointly training the VLM for all tasks, Z-1 enables VLM--Action Expert joint training only when training-stage diagnostics indicate that AE-only optimization is insufficient.
This strategy improves task performance while avoiding unnecessary representation drift and training cost when the SFT representation is already sufficient.

\subsubsection{Effect of Success-Aware Reward Decay}

\begin{figure*}[!t]
    \centering
    \includegraphics[width=\textwidth]{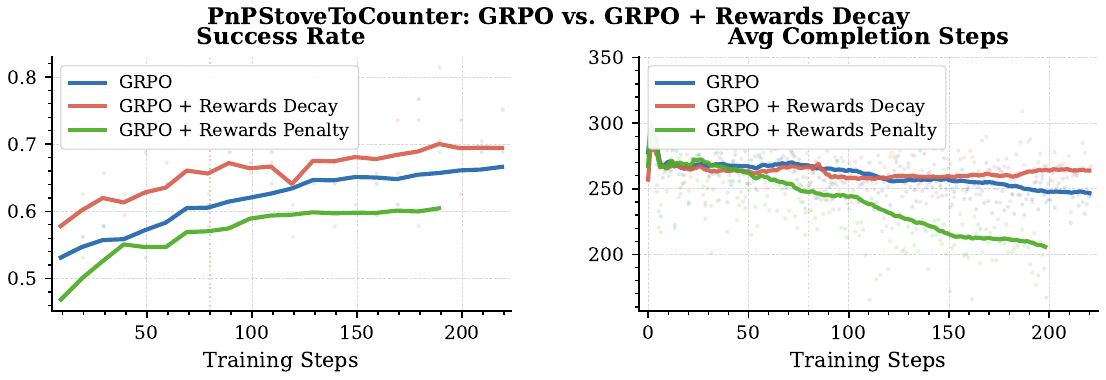}
    \caption{
        Training curves on the PnPStoveToCounter task comparing standard GRPO, GRPO with Reward Penalty, and GRPO with Success-Aware Reward Decay.
        \textit{Left}: success rate over training steps.
        \textit{Right}: average completion steps over training steps, where lower values indicate shorter successful executions.
        Reward Penalty applies the standard mean-normalized GRPO advantage after completion-aware reward calibration, causing slower successful rollouts to receive negative advantages in all-success groups.
        This strongly reduces completion steps but slows success-rate improvement.
        Success-Aware Reward Decay uses a non-negative ordering among successful rollouts, achieving faster success-rate improvement while still encouraging shorter executions.
        Shaded points represent raw values, and solid lines show smoothed trends.
    }
    \label{fig:pnp_stove_to_counter_metrics}
\end{figure*}

We analyze Success-Aware Reward Decay on the PnPStoveToCounter task and compare it with both standard GRPO and a Reward Penalty variant.
The Reward Penalty variant uses the same completion-aware reward calibration but keeps the original mean-normalized GRPO advantage for all groups.
As a result, in all-success groups, successful rollouts that complete later than the group average can receive negative advantages.

As shown in Figure~\ref{fig:pnp_stove_to_counter_metrics}, both reward-based variants improve over standard GRPO, indicating that completion time provides useful information beyond binary task success.
However, the two variants exhibit different optimization behavior.
Reward Penalty substantially reduces average completion steps, showing that explicitly penalizing slower successful trajectories provides a strong signal for shorter executions.
At the same time, its success rate increases more slowly, suggesting that penalizing successful but slower rollouts can make optimization overly aggressive when the policy is still learning reliable task completion.

In contrast, Success-Aware Reward Decay achieves faster and more stable success-rate improvement while also reducing completion steps.
As described in Section~\ref{sec:method}, it replaces mean normalization with a non-negative minimum-baseline ordering in all-success groups, so later successful rollouts are not assigned negative advantages.
Instead, earlier successful completions receive larger positive advantages.
This preserves the sparse success objective while encouraging more direct successful behaviors when they are available.

\section{Conclusion}

We presented Z-1, a reinforcement learning post-training framework for flow-based Vision-Language-Action models.
Built on top of the pretrained $\pi_{0.5}$ policy, Z-1 first adapts to RoboCasa through per-scene supervised fine-tuning using only publicly released demonstrations, and then Z-1 applies a task-wise GRPO post-training strategy selected by training-stage diagnostics.
We introduces a set of complementary modules, including Shared-Prefix GRPO, Tree-Structured Prefix Branching, Success-Aware Reward Decay, and selective VLM--Action Expert joint training, to improve the efficiency, stability, and adaptability of online optimization.

On $24$ RoboCasa manipulation tasks, Z-1 RL improves the average success rate from $67.4\%$ of SFT to $80.6\%$ after the reinforcement post-training.
Z-1 RL achieves the best average performance among the compared published RoboCasa results, establishing a new sota method in this setting.
The improvements are especially large in contact-rich tasks such as drawer manipulation and sink/faucet control, where shared-prefix rollout construction and task-specific GRPO provide a cleaner learning signal for fine manipulation behavior.

Despite these gains, several challenges remain.
Z-1 is less competitive than X-WAM on stove tasks, indicating that long-horizon transport, spatial reasoning, and robust recovery from intermediate failures remain difficult for the current framework.
Future work may further improve Z-1 by incorporating stronger exploration, richer visual-spatial representations, and cross-task RL post-training.
Overall, our results show that carefully designed GRPO post-training can substantially improve flow-based VLA policies without relying on additional private demonstrations.

\newpage

\bibliography{references}
\bibliographystyle{plainnat}

\newpage
\appendix

\section{Full Author List and Contributions}
\textbf{Lang Cao} and \textbf{Renhong Chen} equally contributed to this paper and are responsible for the proposal and implementation of the RL idea.
\textbf{Luyi Li} is responding for the SFT training.
\textbf{Peng Wang} and \textbf{Mofan Peng} are responsible for data processing.
\textbf{Yitong Li} supervised the project and provided guidance on research direction and technical development.

\section{Full Task-wise Results}
\label{app:full_task_results}

Table~\ref{tab:robocasa_full} reports the full task-wise success rates. 
Z-1 RL improves over Z-1 SFT on most tasks, while preserving performance on tasks that are already near saturation after SFT, such as OpenDoubleDoor, CloseDrawer, TurnOffMicrowave, and CoffeePressButton.

\begin{table}[t]
\centering
\caption{
Full task-wise success rates on 24 RoboCasa tasks. 
All numbers are success rates in percentages.
}
\label{tab:robocasa_full}
\resizebox{\textwidth}{!}{
\begin{tabular}{lcccccc}
\toprule
\textbf{Task} 
& \textbf{GR00T} 
& \textbf{GR00T N1.5} 
& \textbf{Video} 
& \textbf{X-WAM} 
& \textbf{Z-1 SFT} 
& \textbf{Z-1 RL} \\
\midrule
OpenSingleDoor        & 59.0 & 62.0 & 68.0 & 96.0 & 95.8 & 95.8 \\
CloseSingleDoor       & 83.0 & 94.0 & 100.0 & 96.0 & 83.3 & 98.4 \\
OpenDoubleDoor        & 15.0 & 89.0 & 96.0 & 94.0 & 100.0 & 100.0 \\
CloseDoubleDoor       & 75.0 & 56.0 & 98.0 & 87.0 & 93.8 & 93.8 \\
OpenDrawer            & 79.0 & 81.0 & 46.0 & 85.0 & 66.7 & 92.2 \\
CloseDrawer           & 99.0 & 100.0 & 96.0 & 100.0 & 100.0 & 100.0 \\
PnPCounterToCabinet   & 36.0 & 57.0 & 42.0 & 67.0 & 54.2 & 76.6 \\
PnPCabinetToCounter   & 20.0 & 26.0 & 36.0 & 73.0 & 37.5 & 76.6 \\
PnPCounterToSink      & 10.0 & 26.0 & 44.0 & 79.0 & 66.7 & 73.4 \\
PnPSinkToCounter      & 33.0 & 46.0 & 64.0 & 71.0 & 58.3 & 79.7 \\
PnPCounterToStove     & 24.0 & 41.0 & 58.0 & 83.0 & 70.8 & 78.1 \\
PnPStoveToCounter     & 29.0 & 60.0 & 64.0 & 80.0 & 70.8 & 84.4 \\
PnPCounterToMicrowave & 13.0 & 33.0 & 52.0 & 62.0 & 39.6 & 50.0 \\
PnPMicrowaveToCounter & 16.0 & 27.0 & 44.0 & 57.0 & 33.3 & 48.4 \\
TurnOnMicrowave       & 78.0 & 66.0 & 92.0 & 82.0 & 89.6 & 89.6 \\
TurnOffMicrowave      & 71.0 & 81.0 & 90.0 & 93.0 & 100.0 & 100.0 \\
TurnOnSinkFaucet      & 63.0 & 85.0 & 84.0 & 92.0 & 29.2 & 98.4 \\
TurnOffSinkFaucet     & 73.0 & 75.0 & 78.0 & 86.0 & 83.3 & 90.6 \\
TurnSinkSpout         & 53.0 & 69.0 & 40.0 & 80.0 & 77.1 & 93.8 \\
TurnOnStove           & 56.0 & 39.0 & 30.0 & 80.0 & 37.5 & 56.2 \\
TurnOffStove          & 27.0 & 21.0 & 6.0 & 35.0 & 20.8 & 32.8 \\
CoffeeSetupMug        & 23.0 & 38.0 & 22.0 & 45.0 & 37.5 & 37.5 \\
CoffeeServeMug        & 73.0 & 67.0 & 76.0 & 82.0 & 79.2 & 95.3 \\
CoffeePressButton     & 85.0 & 94.0 & 92.0 & 96.0 & 92.8 & 92.8 \\
\midrule
\textbf{Average}      & 49.7 & 59.7 & 63.3 & 79.2 & 67.4 & \textbf{80.6} \\
\bottomrule
\end{tabular}
}
\end{table}

\section{Training and Evaluation Details}
\label{app:training_details}

\subsection{SFT Training}
Z-1 uses a two-stage training pipeline consisting of supervised fine-tuning (SFT) followed by task-wise GRPO post-training when selected by training-stage diagnostics. 
The SFT stage is implemented with the official OpenPI JAX training code using $1,199$ publicly released RoboCasa demonstrations. 
During SFT, both the vision-language module and the action expert are updated.

We use scene-specific SFT checkpoints to initialize task-level RL when post-training is applied.
For the door, drawer, and microwave scenes, we adopt a two-stage SFT schedule: the model is first fine-tuned on all $1,199$ RoboCasa demonstrations and then further fine-tuned on the demonstrations from the corresponding scene. 
For the remaining scenes, we directly fine-tune the model on the demonstrations from the corresponding scene. 
This produces one SFT checkpoint pool for each scene category.

\subsection{Checkpoint Selection}
For each scene category, we select a stable, converged SFT checkpoint with held-out validation loss around $0.01$ as the initialization for GRPO post-training.
We do not strictly select the checkpoint with the minimum validation loss, since small validation-loss differences near convergence are not necessarily predictive of downstream rollout performance.
Each RoboCasa task selected for RL then starts from the selected checkpoint of its corresponding scene category.

\subsection{GRPO Post-training}
For GRPO post-training, each selected task is optimized independently from its scene-specific SFT checkpoint. 
Unless otherwise specified, RL updates only the action expert. 
We use AdamW with a learning rate of $5.0 \times 10^{-6}$, weight decay of $0.01$, and gradient clipping of $1.0$. 
The micro-batch size is $16$, the global batch size is $1024$, and the default GRPO group size is $8$. 
Each update uses $4$ rollout epochs with $64$ parallel training environments. 
The maximum episode length is adjusted according to the task. 
We use a clip ratio of $0.2$ and do not use an additional KL regularizer, critic model, or reward model. 
Reward filtering uses lower and upper bounds of $0.1$ and $0.99$, respectively. 
When Success-Aware Reward Decay is enabled, the reward decay factor is set to $0.998$.

\subsection{Ablation-specific Settings}
For the shared-prefix ablation in Figure~\ref{fig:shared_prefix}, we additionally use a simple prefix-decay schedule for the flat SharePrefix variant. 
Starting from training step $4$, the shared prefix length is reduced from $15$ action chunks to $0$ by decreasing one chunk at each subsequent step. 
For the Prefix Tree variant, we use a tree group size of $4$ instead of the default group size of $8$, as this setting led to more stable optimization in preliminary runs.
Because this smaller group size would reduce the number of effective GRPO groups that provide non-zero learning signal under a fixed global batch size, we increase the training batch size for this variant to keep the number of effective learning groups comparable to the default configuration.

\subsection{Evaluation Protocol Details}
\label{app:evaluation_protocol}

We evaluate Z-1 SFT and Z-1 RL with multiple evaluation seeds and independent evaluation rollouts.
For Z-1 SFT, we use $4$ evaluation seeds and run $12$ evaluation rollouts for each seed, resulting in $48$ rollouts per task.
For Z-1 RL, we use $8$ evaluation seeds and run $8$ evaluation rollouts for each seed, resulting in $64$ rollouts per task.
The different numbers of rollouts reflect our evaluation budget at the corresponding training stages; in both cases, task success is estimated from independent rollouts without reusing training trajectories.

The reported success rate for each task is computed over all evaluation rollouts.
Category-level results are computed as the unweighted average of task-level success rates within each category, and the overall average is computed as the unweighted average across all $24$ tasks.

\section{Additional Method Details}
\label{app:additional_method_details}

\subsection{Tree-Structured Prefix Branching Details}
\label{app:tree_prefix_details}

Assume the rollout group size is $G=2^k$. 
Given a full-branching depth $P$ in action chunks, Z-1 defines branching points as
$$
b_{\ell}
=
\left\lfloor
\frac{P}{2^{k-\ell}}
\right\rfloor,
\quad
\ell=1,\ldots,k.
$$
Starting from one shared cluster, each branching point splits every current cluster into two sub-clusters. 
After $k$ splits, all $G$ rollouts continue independently from $b_k=P$.

For partially shared segments, followers execute the corresponding cluster leader's action. 
We store the leader's action, log-probability, and conditioning inputs for all followers in the cluster. 
Under deterministic state cloning and shared policy parameters, these records are on-policy for the follower states. 
The GRPO objective and advantage estimator are unchanged; only the rollout group construction differs.

\subsection{Success-Aware Reward Decay Details}
\label{app:reward_decay_details}

For groups that contain both successful and failed rollouts, we keep the standard group-relative normalization:
$$
A_i
=
\frac{R_i-\mu_R}{\sigma_R+\epsilon}.
$$
This keeps the success-failure distinction as the dominant learning signal.

For groups where all rollouts succeed, standard mean normalization would assign negative advantages to slower but still successful rollouts. 
To avoid turning completion-time calibration into an explicit penalty, we instead use the group minimum as the baseline:
$$
A_i
=
R_i
-
\min_{j} R_j.
$$
This gives the slowest successful rollout zero advantage and gives earlier successful rollouts positive advantages. 
Therefore, Success-Aware Reward Decay does not penalize longer successful trajectories; it only introduces a non-negative ordering among successful rollouts.

\paragraph{Reward Penalty baseline.}
In the reward-decay ablation, we also compare against a Reward Penalty variant.
This variant uses the same completion-aware calibrated returns as Success-Aware Reward Decay but keeps the standard mean-normalized GRPO advantage for all groups, including all-success groups.
Therefore, slower successful rollouts can receive negative advantages when their calibrated returns are below the group mean.

\subsection{Flow-SDE Likelihood Ratio for GRPO}
\label{app:flow_sde_ratio}

Z-1 applies GRPO to a flow-based VLA policy.
A direct application of policy-gradient methods to deterministic flow sampling is non-trivial because the final action chunk is generated through an integration path rather than sampled from an explicit one-step action distribution.
To obtain a tractable policy ratio, we follow the flow-SDE formulation used in recent online RL fine-tuning methods for flow policies~\citep{chen2026pitextttrlonlinerlfinetuning,zhang2026reinflowfinetuningflowmatching}.
The key idea is to convert the deterministic flow generation process into a stochastic Markov process by injecting Gaussian noise into the intermediate flow transitions.
This provides exploration during rollout collection and yields transition densities that can be used for likelihood-ratio policy optimization.

Let $x$ denote the policy conditioning input, including the language instruction, visual observation, and proprioceptive state.
For each action chunk, the flow-based action expert starts from an initial noise variable $z_0$ and generates a sequence of intermediate latent variables
$$
z_0, z_1, \ldots, z_K,
$$
where the final latent $z_K$ is decoded or interpreted as the action chunk $a$.
Under the flow-SDE formulation, each transition is modeled as a Gaussian conditional distribution:
$$
p_{\theta}(z_{k+1} \mid z_k, x)
=
\mathcal{N}
\left(
z_{k+1};
m_{\theta}(z_k, x, t_k),
\sigma_k^2 I
\right),
$$
where $m_{\theta}(z_k, x, t_k)$ is the discretized flow update predicted by the action expert at flow time $t_k$, and $\sigma_k$ is the transition noise scale.
In practice, $m_{\theta}$ is determined by the velocity or denoising prediction of the flow-based action expert and the numerical integration rule used by the policy sampler.

The log-probability of an action chunk is computed as the sum of transition log-probabilities along the sampled flow-SDE path:
$$
\log \pi_{\theta}(a \mid x)
=
\sum_{k=0}^{K-1}
\log p_{\theta}(z_{k+1} \mid z_k, x)
+
C,
$$
where $C$ contains terms independent of $\theta$.
For policy-ratio computation, this constant cancels out.
During rollout collection, we store the sampled flow path together with the rollout-time log-probability under the behavior policy $\pi_{\theta_{\mathrm{old}}}$.
During GRPO optimization, we recompute the log-probability of the same sampled path under the current policy $\pi_{\theta}$ and form the action-chunk ratio as
$$
\rho_{\theta}(x,a)
=
\exp
\left(
\log \pi_{\theta}(a \mid x)
-
\log \pi_{\theta_{\mathrm{old}}}(a \mid x)
\right).
$$

For a trajectory rollout, this ratio is evaluated independently for each trainable action chunk.
Prefix chunks masked by Shared-Prefix GRPO are excluded from the loss and therefore do not contribute to the likelihood-ratio objective.
For Tree-Structured Prefix Branching, only chunks selected as trainable under the branch construction are included.
This design keeps the GRPO update restricted to action chunks whose sampled flow-SDE transitions and rollout-time log-probabilities are recorded.

The resulting objective is the same clipped group-relative objective described in Section~\ref{sec:method}; the flow-SDE formulation only specifies how $\log \pi_{\theta}(a \mid x)$ and $\rho_{\theta}(x,a)$ are computed for flow-based action generation.

\subsection{Implementation of Selective VLM--Action Expert Joint Training}
\label{app:selective_joint_training}

Selective VLM--Action Expert joint training is implemented by changing the trainable parameter set of the OpenPI actor.
It does not change the rollout format, advantage computation, flow-SDE likelihood-ratio computation, or clipped GRPO objective.

In our implementation, this is controlled by the expert-only training configuration of the OpenPI actor.
In the default AE-only setting, expert-only training is enabled, and the PaliGemma vision-language backbone is frozen during GRPO.
The frozen backbone includes the vision-language prefix encoder used to process image and language inputs.
The action expert, which generates flow-based action chunks conditioned on the prefix representation and robot state, remains trainable.

When joint training is enabled, expert-only freezing is disabled, and the vision-language backbone is included in the trainable parameter set together with the action expert.
Thus, the same GRPO loss can update both the perception/language representation and the action-generation module.
This allows task-level reward feedback to adapt visual grounding, language-conditioned spatial representations, and action generation jointly.

During rollout collection, the policy stores the sampled flow-SDE chains, selected denoising indices, rollout-time action log-probabilities, and the corresponding model inputs.
During training, the actor recomputes the current log-probabilities of the stored sampled chains under the updated policy and forms the likelihood ratio against the rollout-time log-probabilities.
The clipped GRPO loss is then backpropagated through the trainable parameters selected for that task.
In AE-only training, gradients are restricted to the action expert.
In joint training, gradients also flow through the vision-language backbone.

The trainable-module configuration is selected using training-stage diagnostics only, such as SFT success rate, early AE-only GRPO progress, and failure modes observed in training rollouts.
The selected configuration is fixed before final evaluation, and no final evaluation rollouts are used for module selection.

\end{document}